\UseRawInputEncoding
\documentclass[conference]{IEEEtran}
\usepackage{cite}
\usepackage{amsmath,graphicx,hyperref,algorithm,algorithmic,enumitem,float}
\usepackage{xcolor,soul}
\usepackage{xurl}
\usepackage[export]{adjustbox}

\IEEEoverridecommandlockouts

\title{When Composition Doesn't Add Up: Humans Identifying Defects in AI-Generated Images}

\author{
\IEEEauthorblockN{Ruoqi Hu$^{1,\dagger}$, Chulin Zhao$^{1,\dagger}$, Jiashuo Chang$^1$, Ramon Ruiz-Dolz$^2$, Hanhe Lin$^{1,2}$}
\IEEEauthorblockA{$^1$Dundee International Institute of Central South University, Central South University, China}
\IEEEauthorblockA{$^2$Faculty of Science, Engineering and Business, University of Dundee, United Kingdom}
\thanks{$^\dagger$ Ruoqi Hu and Chulin Zhao contribute equally.}
}

\begin{document}
%
\maketitle
\begin{abstract}
State-of-the-art text-to-image (T2I) models exhibit pronounced and systematic defects when prompts involve intricate compositional factors such as multiple entities and multiple attributes. In this paper, we investigate how humans identify such defects. 
Specifically, we manually select 651 reference images from the four categories of people, hand, object, and scene that exhibit complex compositional characteristics, from which prompts emphasizing compositional factors are derived by manually editing ChatGPT-generated prompts. We then feed the prompts into three selected T2I models to generate AI images and conduct a comprehensive subjective study to identify their defects. For each image, 29 participants provide multi-label assessments specifying defect types and locations. The study yields the compositional AI-generated image defect (CO-AID) dataset, including reference images, prompts, AI-generated images, and information on defect locations and types. Experimental results show that training a deep model on CO-AID can both predict defects in AI-generated images and optimize AI image generation, demonstrating its usability and effectiveness. 
The database and supplementary materials are available at: \url{https://github.com/Future-IQA/CO-AID}.

\end{abstract}
\begin{IEEEkeywords}
AI-generated image, text-to-image generation, composition, defect identification, quality assessment
\end{IEEEkeywords}

\section{Introduction}
\label{sec:intro}

\begin{figure}[t]
    \centering
\includegraphics[width=\linewidth]{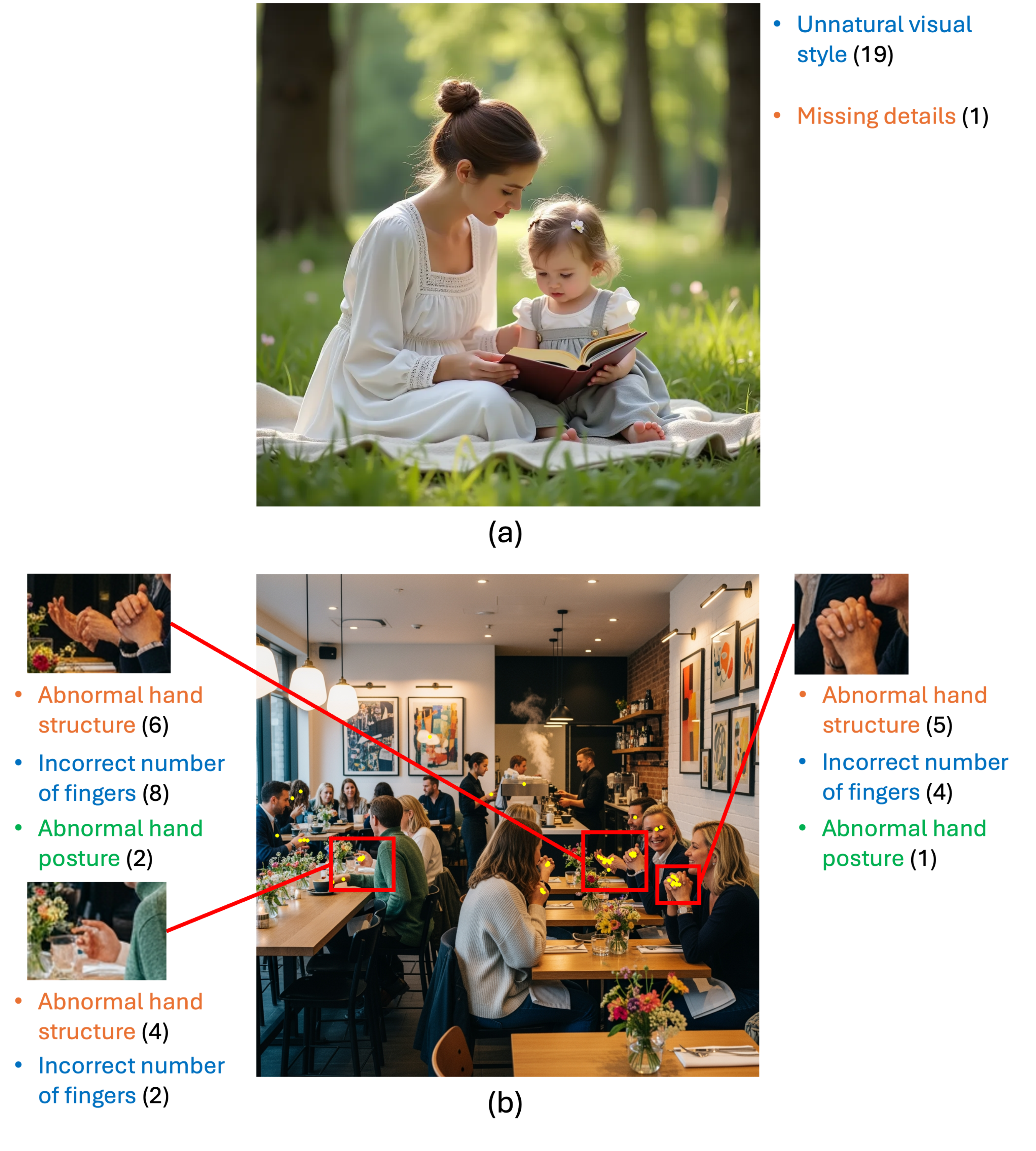}
    \caption{Example AI-generated images in CO-AID dataset. 
    (a) Image with global defect. 19 participants report that it shows a CGI-like appearance. 
    (b) Image with local defects marked by yellow dots. The three most densely clicked regions and their corresponding defect reasons are highlighted.}
    \label{fig:comparison}
\end{figure}

\begin{figure*}[ht]
    \centering
    \includegraphics[width=\textwidth]{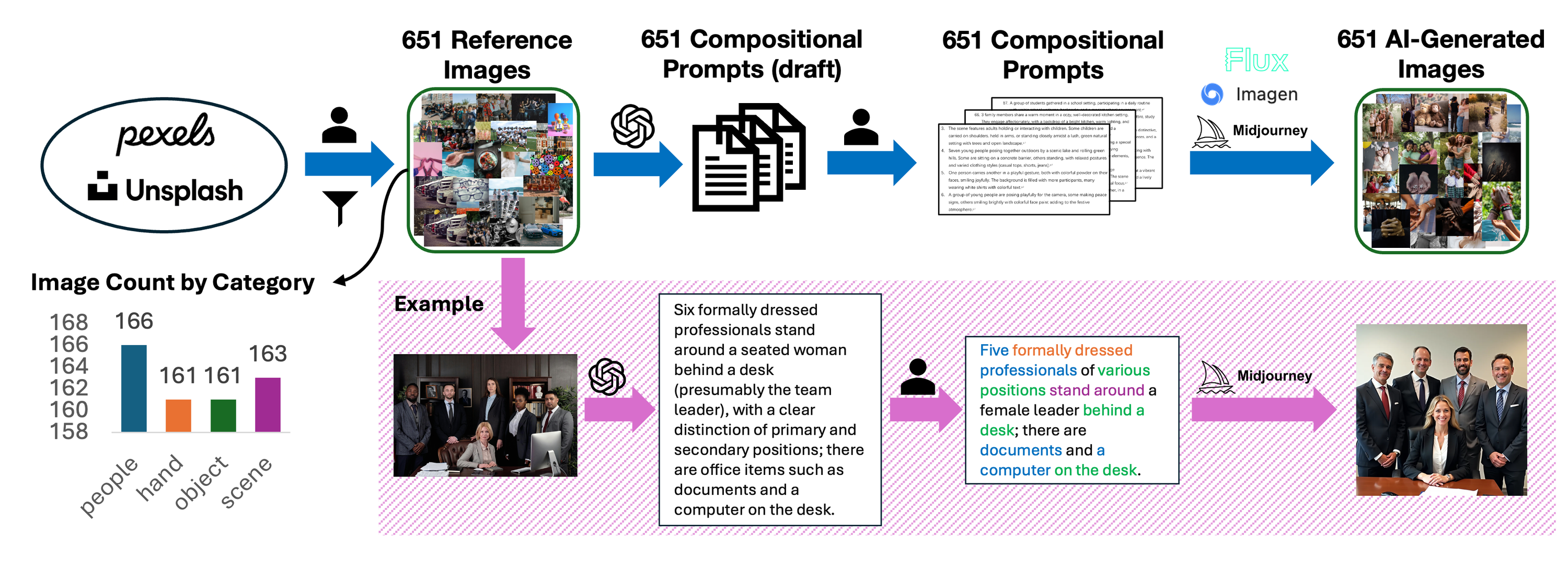}
    \caption{Workflow for collecting the AI-generated compositional images. A total of 651 reference images that exhibit compositional characteristics across four categories of people, hand, object, and scene were manually selected and downloaded from Pexels and Unsplash, which were then used to generate compositional prompts by manually editing ChatGPT-generated prompts. The prompts were then fed into three T2I models Midjourney, Imagen, and Flux to produce AI-generated images for further subjective study.}
    \label{fig:image_workflow}
\end{figure*}

Recently, the development of AI image generation techniques has enabled their widespread application in content creation, artistic design, advertising, virtual reality, and human-computer interaction \cite{ramesh2022hierarchical}. 
Existing research on AI image generation can primarily be categorized into tasks such as image inpainting~\cite{yu2019free}, image editing~\cite{brooks2023instructpix2pix}, image-to-image translation~\cite{isola2017image}, and text-to-image (T2I) generation~\cite{saharia2022photorealistic}. 
Among these tasks, T2I generation is most challenging, as it requires cross-modal translation from language to vision, fine-grained alignment between textual elements and visual content, and effective representation of heterogeneous modalities within a unified generative framework~\cite{baltruvsaitis2018multimodal}.

The development of T2I models underwent a critical technological transition from GAN-based models \cite{xu2018attngan} to diffusion-based models~\cite{saharia2022photorealistic}, leading to significant improvements in visual quality and semantic consistency~\cite{dhariwal2021diffusion}. Recent diffusion-based models, such as Stable Diffusion \cite{rombach2022high}, are good at generating images with high resolution, rich details, and strong semantic alignment from textual prompts, provided the prompts focus on a single subject or a single attribute, such as depicting a person`s appearance or clothing. However, when the generation scenario shifts to complex views, their performance tends to degrade significantly. Further analysis reveals that this degradation is closely related to the compositional factors embedded in the prompts~\cite{hu2023tifa}. 


Composition has been widely recognized as a key factor influencing the performance of T2I generation \cite{hu2023tifa,huang2025t2i}. Existing research on compositional T2I generation focuses primarily on faithfulness scores and subjective assessments. The former focuses on whether generated images satisfy the compositional factors specified in the text. Representative approaches generate controlled compositional prompts or decompose prompts into task-level probes (e.g., counting, attribute binding, and object relations). The faithfulness scores are obtained using automated, task-level metrics to evaluate model output \cite{huang2025t2i,hu2023tifa}. The latter evaluates compositional T2I generation through subjective assessments. Petsiuk \textit{et al.}~\cite{petsiuk2022human} introduced a multi-task benchmark with varying levels of compositional difficulty, in which model performance is evaluated globally by human observers. Saharia \textit{et al.}~\cite{saharia2022photorealistic} collected human preference ratings across various prompt categories that include compositional scenarios, allowing subjective performance comparison of models. Despite these efforts, existing compositional benchmarks focus on global evaluation and lack fine-grained local assessment. 

Several studies have investigated the identification of local defects in AI-generated images. For example, Fang \textit{et al.}~\cite{fang2024humanrefiner} assessed the quality of AI-generated images by collecting clicks of artefact regions, feedback of prompt-image mismatch, and perceptual scores. Liang \textit{et al.}~\cite{liang2024rich} constructed the AbHuman benchmark, in which bounding-box and categorized annotations for abnormal human body parts such as distorted hands and limbs were collected from the human observers. However, the AI-generated images considered in these studies are relatively simple and ignore compositional factors. Consequently, the ways in which humans precisely localize, perceive, and explain compositional defects within complex AI-generated images remain underexplored.

There is no explicit definition of \textit{composition} in the context of AI-generated images. Building on existing research~\cite{huang2025t2i,hu2023tifa}, we define a highly compositional AI-generated image as \textit{an image showing a complex view generated from a compositional prompt that simultaneously involves multiple entities, multiple attributes, multi-dimensional spatial relations, and complex interactions}. 

To investigate how humans identify compositional defects in AI-generated images, we conduct a subjective study. Based on this definition of composition, we generate 651 compositional prompts with reference to real images from categories of people, hand, object, and scene and feed them into three selected T2I models (Midjourney~\cite{midjourney}, Imagen~\cite{imagen}, and FLUX~\cite{flux}) to generate compositional image samples. We design a rigorous framework to conduct a subjective study on the sampled images, where 29 participants identify compositional defects and types. Examples of images and their corresponding annotations are shown in Fig.~\ref{fig:comparison}.

The main contributions of this paper are as follows.
\begin{itemize}
    \item We propose a framework for identifying compositional defects in AI-generated images. To the best of our knowledge, this is the first work to systematically study this topic. 
    
    \item We design and conduct a lab-based subjective study on 651 AI-generated images assessed by 29 participants, and curate a novel dataset named the COmpositional AI-generated Image Defect (CO-AID) dataset. 

    \item We perform a comprehensive result analysis of the three T2I models. Proof-of-concept experiments demonstrate that training a deep model on CO-AID enables both the prediction of defects in AI-generated images and the optimization of AI image generation.

\end{itemize}

\begin{figure*}[t]
    \centering
    \includegraphics[width=\textwidth]{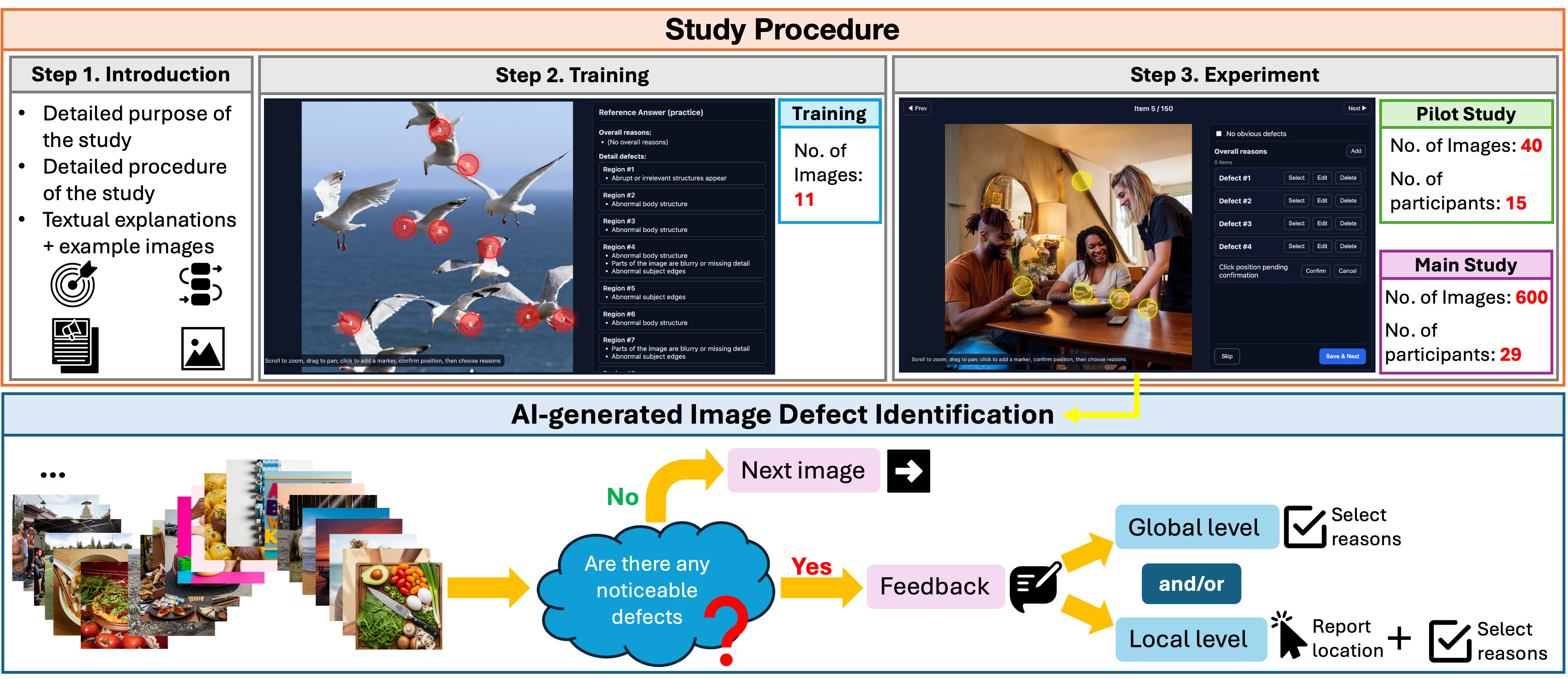}
    \caption{Overview of our proposed human perceptual defect identification study. The study procedure consists of three steps: Introduction, Training and Experiment. Before the main study, a small pilot study was conducted to improve experimental instruction, complete and polish optional defect reasons, and optimize the graphical user interface. In the experiment step, participants sequentially view AI-generated compositional images and identify their defects. For those who show noticeable defects, participants provided feedback at the global and/or local levels.}
    \label{fig:workflow}
\end{figure*}

\section{AI-generated Compositional Image Collection}
\label{sec:data preparation}

Fig.~\ref{fig:image_workflow} shows the procedures of AI-generated compositional image collection. A total of 651 reference images exhibiting complex compositional characteristics were manually selected from two image websites (Pixels and Unsplash), spanning the categories of people, hand, object, and scene. Next, 651 prompts emphasizing compositional factors were obtained by manually editing ChatGPT-generated prompts derived from the 651 reference images. Lastly, 651 AI images were generated by feeding the prompts into Midjourney, Imagen, and FLUX models.


\subsection{Reference Image Selection}


We aim to collect a set of AI-generated compositional images with content diversity while avoiding the generation of images that are merely figments of imagination. We found that there were four major categories that are highly likely to exhibit compositional characteristics, including people, hand, object, and scene. Here, `hand' is treated as an independent category, as it typically exhibits high structural complexity and rich interaction semantics, which are particularly challenging for T2I models~\cite{cai2016understanding, fang2024humanrefiner}. Moreover, to cover a wide range of entity types, we further divided the category of object into sub-categories of animal, machine, LEGO, food, and grocery. To cover a wide range of spatial contexts and visual structures, we further divided the category of scene into modern, natural, and abstract (pattern and text). 
Eventually, we manually selected and downloaded 651 reference images with free-use licenses from the Pexels~\cite{pexels} and Unsplash~\cite{unsplash} websites using their official application program interfaces (APIs). The numbers of images in the categories of people, hand, object, and scene are 166, 161, 161, and 163, respectively.

\subsection{Compositional Prompt Generation}

To generate compositional prompts from 651 reference images, we first fed them into ChatGPT to generate textual descriptions, on which manual refinement and modification were conducted to ensure that the generated prompts meet the following criteria:
(1) explicitly describe multiple entities and their attribute information, such as appearance, posture, and state; (2) explicitly explain two-dimensional or three-dimensional spatial relations among entities, such as relative positions and occlusion relationships; and (3) explicitly outline interaction between entities and/or their environments, such as physical contact and hand-object interaction. 
Through this process, we ultimately obtained a set of 651 compositional prompts that exhibit clear compositional characteristics at the semantic level. 

\subsection{Compositional Image Generation}

Given the constructed compositional prompts, we further generated the corresponding compositional images using the state-of-the-art T2I models.

We conducted a preliminary study to identify suitable T2I models for our image generation task. Specifically, we considered several T2I models known for their strong generative capabilities, including those in academia, namely Stable Diffusion~\cite{rombach2022high}, Imagen~\cite{saharia2022photorealistic}, and DALL$\cdot$E~\cite{ramesh2022hierarchical}, and those in industry, namely Midjourney~\cite{midjourney}, Firefly~\cite{firefly}, and FLUX~\cite{flux}. We randomly sampled 80 compositional prompts and fed them into those models to generate AI images. By perceptual comparison of the generated results, we found that Midjourney, Imagen, and FLUX generate higher-quality compositional images and therefore selected them for our study. 

The 651 compositional prompts were also randomly divided into three groups, each group assigned to one of the selected T2I models for AI image generation. The 651 AI-generated images were divided into three groups of 11, 40, and 600 images, which were used for training, pilot study, and main study, respectively, as detailed in the next section.

\section{Human Perceptual Defect Identification Study}
\label{sec:human_perceptual_quality_assessment}

Fig.~\ref{fig:workflow} illustrates the overview of our proposed study on human perceptual identification of defects in AI-generated images. The study consists of three steps: introduction, training, and experiment. 
In this study, participants were required to conduct a careful visual inspection and provide feedback on defects at both the global and local levels. Before the main study, a pilot study was conducted to improve experimental instruction, complete and polish optional defect reasons, and optimize the graphical user interface.

\subsection{Study Procedure}

\begin{description}
\item[Step 1.~Introduction] After being informed about data protection practices and signing a consent form, participants were presented with a detailed introduction to the purpose and procedure of the study, together with a combination of textual explanations and example images.

\item[Step 2.~Training] Participants began the training after reviewing the instructions. 11 images were selected to familiarize participants with study procedure. After identifying defects for each image, the suggested defects marked by our two researchers were displayed as a reference. Considering the difference in human visual experience, participants were not required to conform their responses to the suggested defects.

\item[Step 3.~Experiment] After completing training, participants proceeded to the formal experiment and provided comprehensive feedback on compositional defects in AI-generated images. The method for AI-generated image defect identification is detailed in the following subsection. 

\end{description}

 \subsection{AI-generated Image Defect Identification}

For each AI-generated compositional image, participants were asked to carefully inspect it. If there was no noticeable defect in an image, participants selected the option of `No noticeable defect' and proceeded to the next image. Otherwise, participants provided feedback on defects at global and/or local levels. 

When defects were difficult to precisely locate or when numerous severe defects made individual annotation impractical, participants report defects at the global level. For images with only a few defects, participants reported defect locations by clicking on the image and selecting the corresponding defect reasons. In addition, participants were allowed to type a description of an observed defect if the defect reason was not listed. Note that for each image, participants were free to provide global-level feedback, local-level feedback, or both.


Based on previous research, we listed four global-level defect reasons: unnatural visual style~\cite{nightingale2022ai}, severe blurriness or missing details~\cite{nightingale2022ai}, large-scale abnormal texts~\cite{bosheah2025challenges}, and violations of commonsense reasoning~\cite{kamali2024distinguish} (e.g., most people wear heavy winter clothing on a summer beach). We also included three global-level defect reasons that are highly relevant to composition~\cite{hu2023tifa,huang2025t2i}, including a large number of abnormal entities (e.g., most entities exhibit attribute binding failures in different aspects such as structure, scale, and color), a large number of anomalous inter-entity interactions (e.g., many abnormal intersecting limbs in multi-person physical contact), and a large number of spatial relationship anomalies (e.g., many illogical perspective or occlusion relationships).

We grouped local-level defects into five categories, including face, hair/fur, hand, body, and object. The grouping scheme shows broad generality and can accommodate entity instances in four different content categories. For example, a face may refer to a human face, an animal face, or a face of a non-realistic entity such as a doll or a LEGO figure. Similarly, hair/fur applies to both human hair and animal fur. For each category, we listed common defect reasons. For example, the defect reasons of a hand include abnormal hand structure, incorrect number of fingers, abnormal hand posture, and abnormal nail structure.



\begin{figure}[t]
    \centering
     \includegraphics[width=\linewidth]{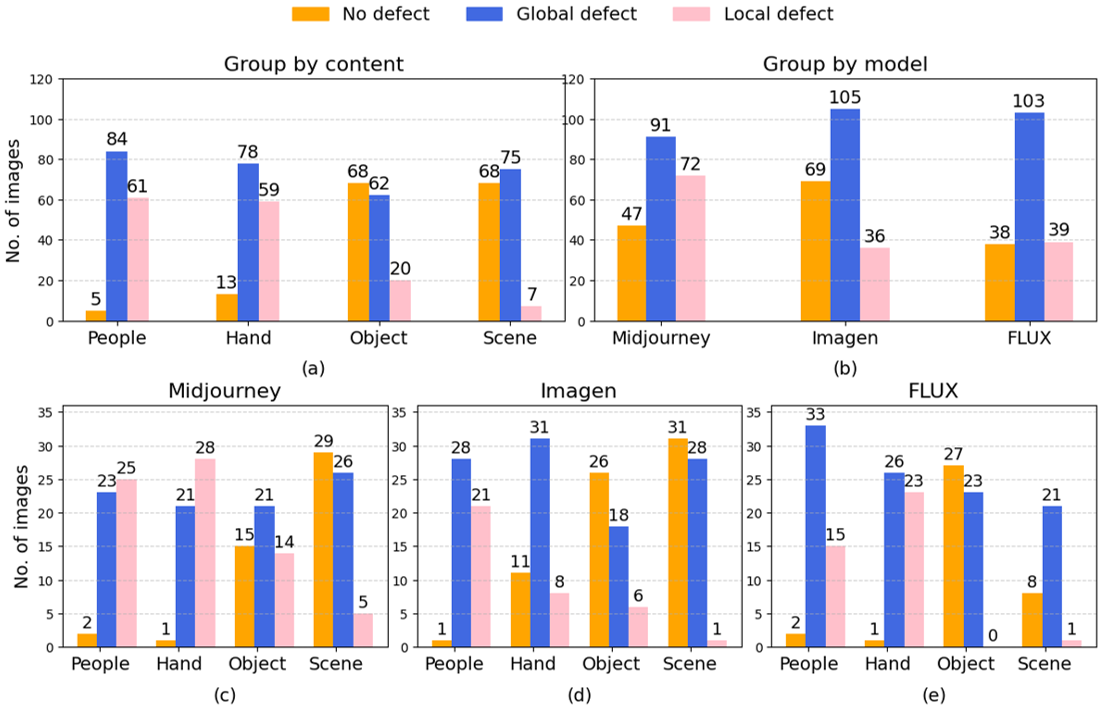}
     \caption{Statistics of the experimental results.}
     \label{fig:overall}
\vspace{-10pt}
\end{figure} 

\section{Experiment}
\label{sec:typestyle}


\subsection{Experimental Setup}

40 images were used for the pilot study and 600 images were used for the main study. 15 and 29 participants were recruited in the pilot and main studies, respectively. To mitigate the effect of fatigue caused by prolonged tasks, the 600 images in the main study were divided into four batches, participants were allowed to take a break after completing a batch. To verify the reliability of the participants, 12 selected images were repeated in the study, resulting in 153 images in each batch. In addition, we recorded the experiment time for further analysis. The entire experiment lasted two weeks and each participant spent approximately 10 hours to complete the study. In total, 18,906 defect annotations were collected, in which the numbers of global level annotations and local level annotations are 6,626 and 12,280, respectively.


\begin{figure*}[t]
\centering
\setlength{\tabcolsep}{2pt}
\renewcommand{\arraystretch}{1.0}
\includegraphics[width=1.4\linewidth,height=0.58\textheight,keepaspectratio]{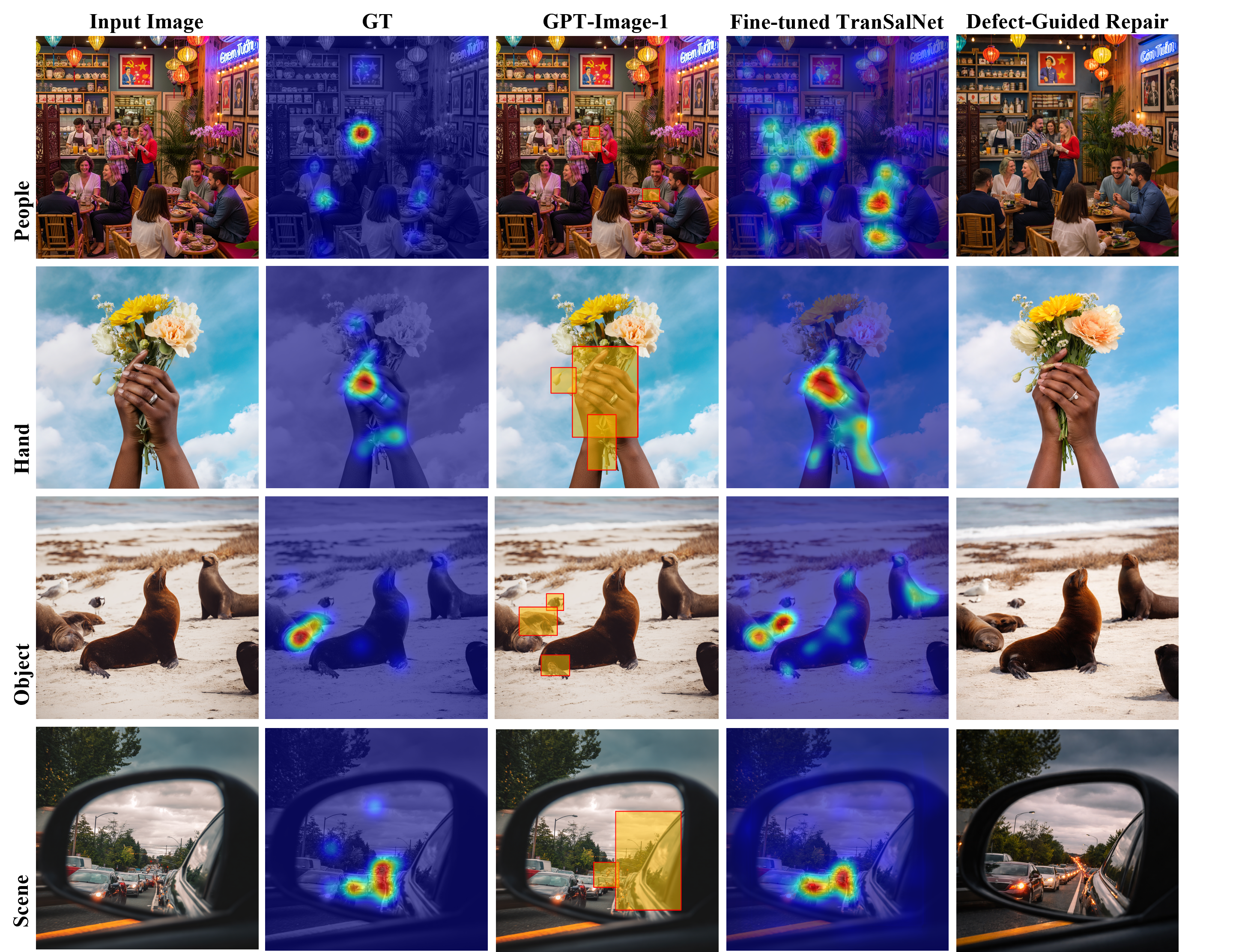}
\caption{Visualization of the proof-of-concept experimental results. From left to right, the figure shows the input AI-generated images, human-annotated ground truth (GT) defect maps, defect detection results (bounding boxes) by GPT-Image-1, defect detection results by fine-tuned TranSalNet model~\cite{lou2022transalnet}, and the corresponding defect-guided image repair results. Zoom in for more details. }
\label{fig:poc_grid}
\vspace{-5pt}
\end{figure*}

\subsection{Outlier Removal}


To ensure the reliability of collected data, we detected and removed outliers. 
We used three criteria to evaluate the reliability of the participants, including engagement, self-consistency, and conformity. Engagement was measured by the time a participant spent annotating each image, with a longer annotation time indicating greater engagement. Self-consistency was assessed by estimating the agreement of a participant’s annotations on repeated images. Conformity was measured by the degree of agreement between the annotations of a participant and those of the majority.

Participants who did not pass the defined thresholds of these criteria were marked as outliers. In total, 6 participants were identified as outliers and the data from the rest 23 participants were used for subsequent analysis. Detailed procedures for outlier detection and removal are provided in the supplementary materials.

\subsection{Result Analysis}

We grouped the 600 images into three categories based on the feedback of 23 participants, including No defect, Global defect, and Local defect. Images labeled as No defect correspond to those for which most of the participants reported no defect. Similarly, image labeled as Global defect and Local defect corresponds to those for which most of the participants reported global-level and local-level defects, respectively.


Fig.~\ref{fig:overall} shows the statistics of the experimental results. As shown in Fig.~\ref{fig:overall}(a), the total number of defect-free images containing people and hands is 18, which is much lower than the 136 images depicting objects and scenes, indicating that generating defect-free images with people and hands is still challenging. 
Among the three T2I models, Imagen outperforms Midjourney and FLUX, generating nearly 70 defect-free images (Fig.~\ref{fig:overall}(b)). However, feedback from participants reports that Imagen is more likely to generate unrealistic and CG-style images, resulting in half of its generated images being labeled as Global defect. In contrast, while Midjourney is more likely to generate realistic images, they often show noticeable detail issues, with 72 images labeled as Local defect. FLUX performs poorly on the image categories of people, hand, and scene, except for the one of object.

\subsection{Proof-of-concept Experiment}
To demonstrate the effectiveness of CO-AID, we conducted a proof-of-concept experiment. We used the locations of human-annotated local-level defects and generated ground truth defect maps. Given the input AI-generated images and their corresponding defect maps, we fine-tuned the TranSalNet~\cite{lou2022transalnet} model that is trained for attention saliency prediction. For comparison purposes, we also generated the defect prediction results using OpenAI's GPT-Image-1 model~\cite{openai}. In addition, we further fed the AI-generated images and their defect maps predicted by the fine-tuned TranSalNet into the GPT-image-1 model for defect-guided image repair, where the defect maps serve as spatial guidance indicating areas that need to be corrected.


Fig.~\ref{fig:poc_grid} shows the experimental results of four sampled images. As can be seen, compared to the GPT-Image-1 model (e.g., hand), the fine-tuned TranSalNet model produces more human-like defect localization patterns, demonstrating the usability of our collected data. Moreover, the repaired images guided by the predicted defect maps correct the defects and therefore show noticeable quality improvement. The experiment demonstrates that our CO-AID dataset can help both predict defects in AI-generated images and optimize AI image generation.

\section{Conclusion}
\label{sec:Conclusion}

We conducted a first-of-its-kind subjective study to systematically identify and categorize defects in AI-generated images. Through a preliminary yet comprehensive human study, we curated the CO-AID dataset and analyzed defect patterns in compositional image generation. Our findings reveal that generating defect-free images involving complex people and hands remains a significant challenge for state-of-the-art T2I models. Furthermore, our proof-of-concept experiments demonstrate the usability and effectiveness of CO-AID as a benchmark for defect identification. Future work will focus on expanding CO-AID into a large-scale benchmark dataset and developing a more generalized defect prediction model. We also aim to explore reward-model-based training strategies to help T2I models generate more realistic, coherent, and defect-free compositional images.

{\small
\bibliographystyle{IEEEbib}
\bibliography{strings,refs}
}

\end{document}